\documentclass[10pt,twocolumn,letterpaper]{article}

\usepackage[pagenumbers]{cvpr} 
\usepackage{graphicx}
\usepackage{amsmath}
\usepackage{amssymb}
\usepackage{booktabs}
\usepackage{amsfonts}
\usepackage{bbding} 
\usepackage{color}
\usepackage{multirow}
\usepackage{multicol}
\usepackage[accsupp]{axessibility}
\usepackage{makecell}
\usepackage{colortbl} 
\usepackage{caption}
\definecolor{Gray}{gray}{0.92} 

\usepackage{graphics} 
\usepackage{epsfig} 
\usepackage{mathptmx} 
\usepackage{times} 
\usepackage{amsmath} 
\usepackage{amssymb}  
\usepackage{siunitx}  
\usepackage{newtxtext,newtxmath} 

\usepackage{times}  
\usepackage{helvet}  
\usepackage{courier}  
\usepackage[hyphens]{url}  
\usepackage{graphicx} 
\usepackage{caption} 
\usepackage{amsmath}
\usepackage{amssymb}
\usepackage{colortbl}
\usepackage{subcaption}
\usepackage{booktabs} 
\usepackage{multirow}
\usepackage[dvipsnames, svgnames, x11names]{xcolor}
\definecolor{lgray}{gray}{0.7}
\usepackage{algorithm}
\usepackage{algorithmic}
\usepackage{enumitem}
\usepackage{bibentry}

\usepackage{newfloat}
\usepackage{listings}

\definecolor{cvprblue}{rgb}{0.21,0.49,0.74}

\usepackage[pagebackref,breaklinks,colorlinks]{hyperref}
\usepackage[capitalize]{cleveref}
\crefname{section}{Sec.}{Secs.}
\Crefname{section}{Section}{Sections}
\Crefname{table}{Table}{Tables}
\crefname{table}{Tab.}{Tabs.}
 
\def\confName{CVPR}
\def\confYear{2024}

\DeclareUnicodeCharacter{0301}{\'{e}}

\begin{document}

\title{FVAR: Visual Autoregressive Modeling via Next Focus Prediction}
\author{
Xiaofan Li\textsuperscript{*},
Chenming Wu\textsuperscript{*},
Yanpeng Sun,
Jiaming Zhou,
Delin Qu, \\
Yansong Qu, 
Weihao Bo,
Haibao Yu,
Dingkang Liang \\
\vspace{4mm}
\textsuperscript{} Baidu Inc.\\
}

\maketitle

\renewcommand{\thefootnote}{\fnsymbol{footnote}}
\footnotetext[1]{Equal contribution.}
\begin{abstract}
Visual autoregressive models achieve remarkable generation quality through next-scale predictions across multi-scale token pyramids. However, the conventional method uses uniform scale downsampling to build these pyramids, leading to aliasing artifacts that compromise fine details and introduce unwanted jaggies and moiré patterns. To tackle this issue, we present \textbf{FVAR}, which reframes the paradigm from \emph{next-scale prediction} to \emph{next-focus prediction}, mimicking the natural process of camera focusing from blur to clarity. Our approach introduces three key innovations: \textbf{1) Next-Focus Prediction Paradigm} that transforms multi-scale autoregression by progressively reducing blur rather than simply downsampling; \textbf{2) Progressive Refocusing Pyramid Construction} that uses physics-consistent defocus kernels to build clean, alias-free multi-scale representations; and \textbf{3) High-Frequency Residual Learning} that employs a specialized residual teacher network to effectively incorporate alias information during training while maintaining deployment simplicity. Specifically, we construct optical low-pass views using defocus point spread function (PSF) kernels with decreasing radius, creating smooth blur-to-clarity transitions that eliminate aliasing at its source. To further enhance detail generation, we introduce a High-Frequency Residual Teacher that learns from both clean structure and alias residuals, distilling this knowledge to a vanilla VAR deployment network for seamless inference. Extensive experiments on ImageNet demonstrate that FVAR substantially reduces aliasing artifacts, improves fine detail preservation, and enhances text readability, achieving superior performance with perfect compatibility to existing VAR frameworks.

\end{abstract}

\begin{figure}[t]
\centering
\includegraphics[width=\linewidth]{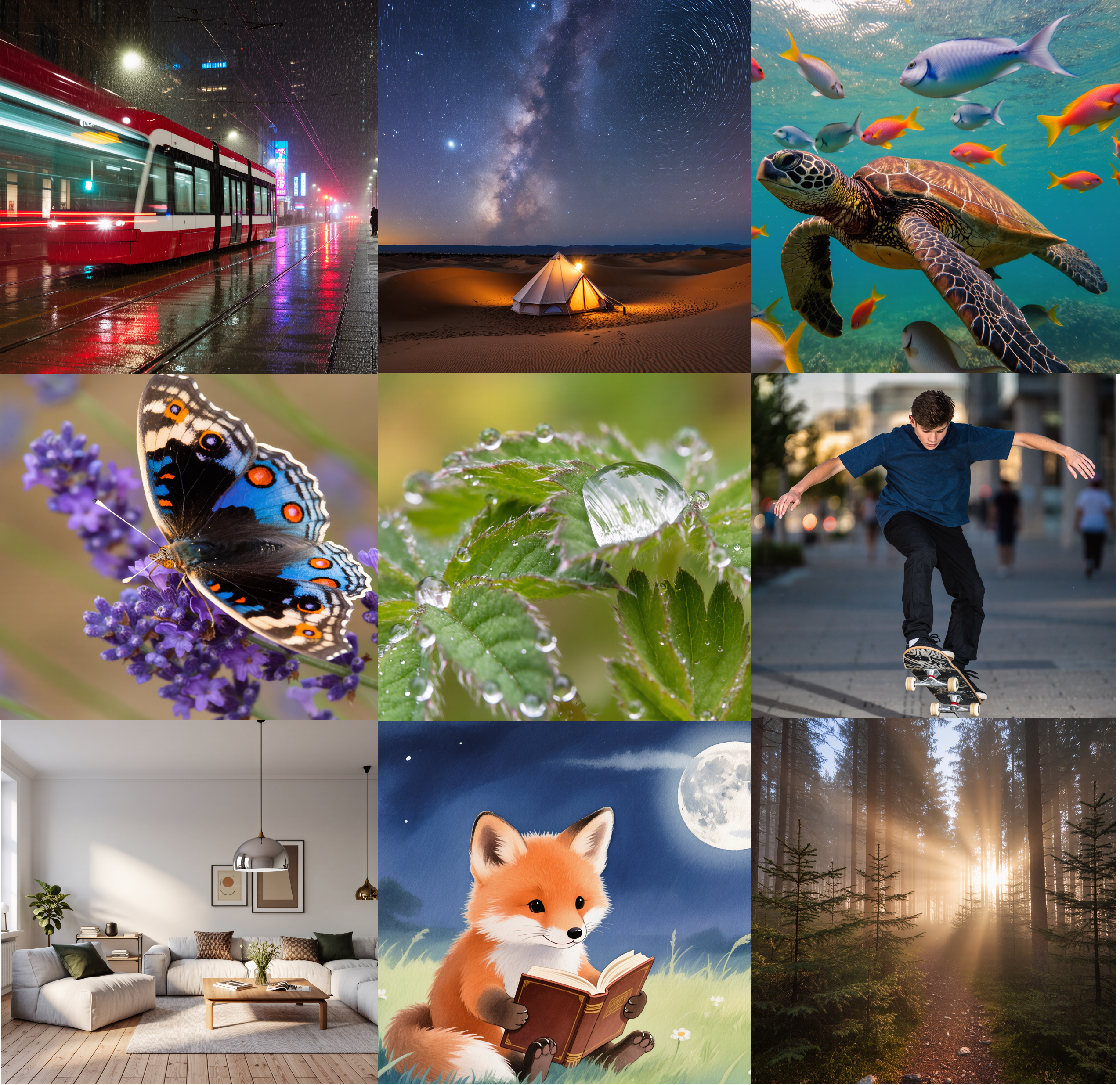}
\caption{\textbf{FVAR achieves superior image generation quality.} Our method generates images with significantly reduced aliasing artifacts (jaggies, moiré patterns) while preserving fine details and text readability compared to standard VAR. The progressive refocusing paradigm enables clean multi-scale representations that lead to sharper, more realistic results. Results shown here are from models trained on additional large-scale datasets (see supplementary material for details).}
\vspace{-2.em}
\label{fig:teaser}
\end{figure}

\section{Introduction}
Large language models (LLMs) commonly operate under the autoregressive paradigm, showcasing impressive scalability and generalizability in both understanding and generating discrete text. This success has spurred research into autoregressive generation across other data modalities. For example, visual autoregressive often relies on quantization-based methods~\citep{vqvae,vqvae2,vqgan,yu2024language} to encode the data into a discrete space. The recently proposed VAR~\cite{var} demonstrates strong scalability and performance compared to diffusion models by structurally predicting from coarse to fine resolutions.

Discrete visual representations based on vector quantization enable autoregressive generation, but they introduce a major challenge: information loss from quantization. Such loss degrades the reconstruction fidelity of discrete image tokenizers and consequently limits overall generation quality~\cite{ldm}. In addition, discrete representations can impair the model’s perception of fine-grained, low-level details, reducing its ability to capture continuous variations and subtle nuances.
Recent work has pursued multiple directions to mitigate these issues. Stronger tokenizers, such as LlamaGen~\cite{llamagen} and ViTVQ~\cite{vit-vqgan}, aim to improve the expressiveness of discrete codes. Continuous autoregressive approaches seek to address quantization errors through strictly proper scoring rules~\citep{tschannen2023givt} or diffusion-based token generation~\citep{li2024autoregressive}. Meanwhile, computationally efficient designs like M-VAR decouple intra- and inter-scale dependencies using linear state-space modules.

However, all of the aforementioned approaches fundamentally rely on uniform downsampling to construct multi-scale pyramids, disregarding the physical principles of optical image formation. As a result, high-frequency content above the Nyquist limit is folded into the baseband, producing aliasing artifacts such as jagged edges, staircasing, and moiré patterns. Consequently, the autoregressive Transformer is forced to simultaneously de-alias these artifacts and generate fine details, which can destabilize training—especially for images containing regular textures or small fonts.

Our work draws inspiration from the physical optics of camera focusing, which reframes visual autoregression from \emph{next-scale prediction} to \emph{next-focus prediction}. The key insight behind our proposed FVAR is that image formation progresses naturally from blur to sharpness through the focusing process, rather than through uniform downsampling that introduces aliasing. Instead of predicting a coarser scale via lossy resolution reduction, we predict the next focus state by gradually decreasing optical blur. This shift in perspective allows us to construct multi-scale representations that are physically grounded and inherently free of aliasing artifacts.

Building on the next-focus prediction paradigm, our method introduces three key technical contributions. \textbf{First}, we construct progressive refocusing pyramids using physics-consistent defocus kernels with gradually decreasing radii, enabling smooth transitions from blur to sharpness and eliminating aliasing at its origin. \textbf{Second}, to enrich detail generation beyond what optical low-pass filtering can provide, we propose a dual-path design that jointly models clean structural content and high-frequency residuals. \textbf{Third}, we develop a High-Frequency Residual Teacher, which learns to integrate these complementary signals during training while preserving full compatibility with a vanilla VAR architecture at inference time. This ensures that the benefits of alias-aware learning are retained without requiring any architectural changes at deployment. As shown in \cref{fig:teaser}, FVAR substantially improves generation quality by reducing artifacts and better preserving fine details.
The technical contributions of our work are summarized as follows:

\begin{itemize}
  \item We reframe visual autoregression from next-scale prediction to next-focus prediction, shifting the core paradigm from uniform downsampling to progressive optical refocusing that mirrors the natural camera focusing process.
  \item We introduce a physics-consistent progressive refocusing pyramid built with defocus kernels of decreasing radius, enabling smooth transitions from blur to sharpness and inherently eliminating aliasing artifacts at their source.
  \item To further improve detail generation, we adopt a dual-path high-frequency residual learning strategy with a High-Frequency Residual Teacher, which learns to exploit both clean structure and residual signals during training and distills this knowledge into a vanilla VAR model for deployment. The resulting design boosts generation quality with zero inference overhead and remains fully compatible with existing VAR frameworks.
\end{itemize}

\section{Related Work}
\begin{figure*}[t]
\centering
\includegraphics[width=0.8\textwidth]{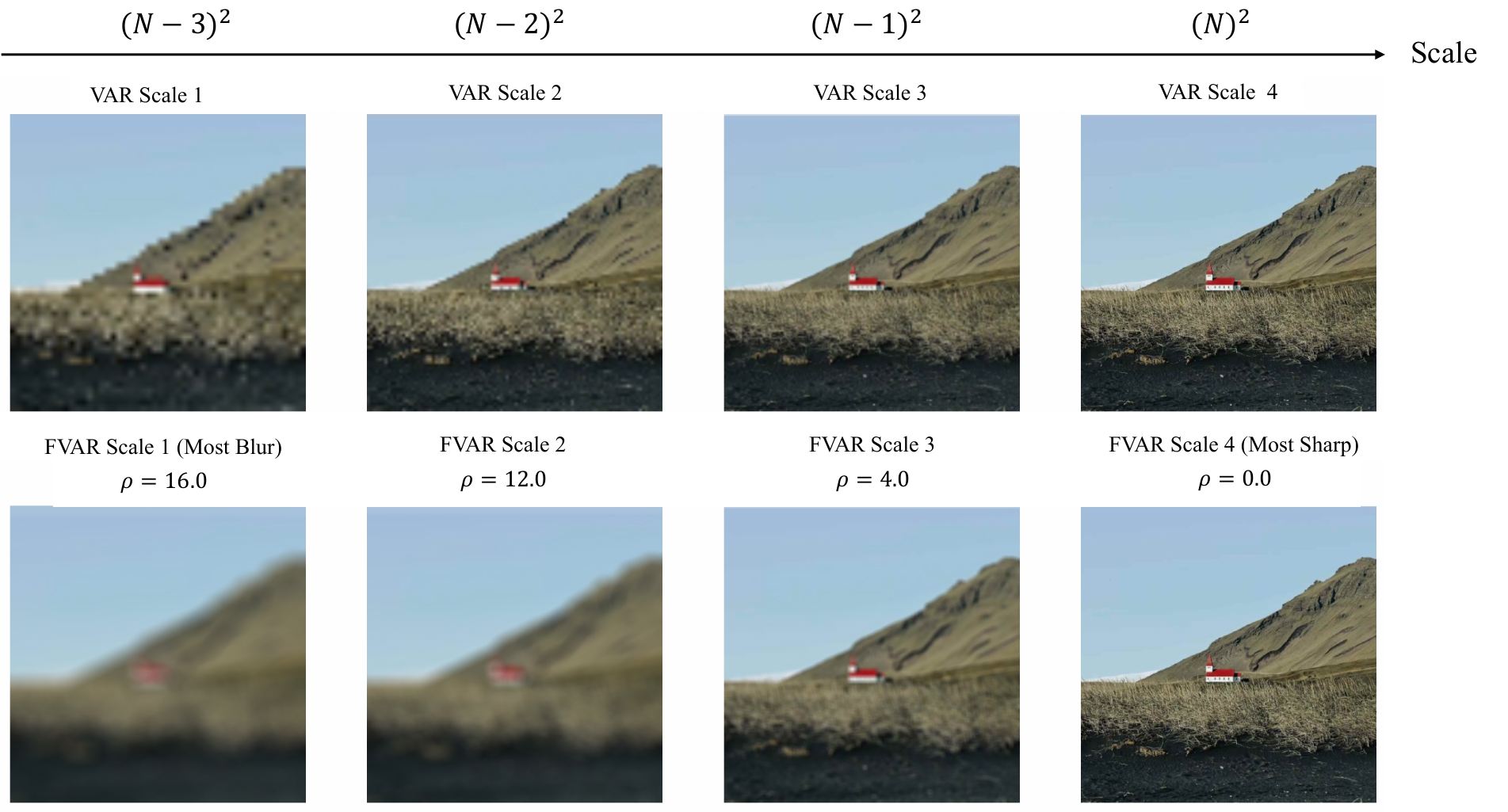}
\caption{\textbf{Progressive Refocusing vs. Uniform Downsampling.} Our method shifts the paradigm from ``next-scale prediction'' to ``next-focus prediction.'' (Left) Standard VAR uses uniform downsampling, introducing aliasing artifacts from coarse to fine scales. (Right) Our proposed FVAR employs progressive refocusing with decreasing PSF radius, mimicking camera focusing from blur to clarity. This physics-consistent approach eliminates aliasing at the source while preserving fine details through dual-path tokenization.}
\label{fig:method}
\vspace{-1.5em}
\end{figure*}
\subsection{Visual Autoregressive Generation}
Visual generation is a rapidly evolving field that integrates ideas from natural language processing with advances in Transformer architectures. Autoregressive (AR) models, inspired by successes in language modeling, tokenize images into discrete codebook indices and predict sequences of tokens in a GPT-style manner \citep{Xiong2024Survey, Goodfellow2014, Brock2018}. Masked-prediction models, similar to BERT, are applied to the visual domain, learning to predict masked tokens within images \citep{Devlin2018, maskgit}, and have been extended to video generation \citep{Yu2023}. Parallelly, continuous diffusion models approach generation through a probabilistic denoising process and have demonstrated state-of-the-art image quality across various datasets \citep{adm, Ho2020}. GANs, the first deep learning-based visual generators, demonstrate compelling fidelity and diversity in generated images \citep{Goodfellow2014, Brock2018, Karras2019}, though they may struggle with distribution coverage and training stability. Likelihood-based generators, including diffusion models, masked prediction, and AR, mitigate these limitations by optimizing tractable objectives or surrogates. In AR pipelines, image tokenization (e.g., VQ-VAE/VQGAN) converts pixels into semantic token spaces to facilitate sequence modeling \citep{Kingma2013, vqvae}. However, discretization introduces quantization ceilings that may compromise fine-detail fidelity. In this context, FVAR builds upon the strengths of autoregressive models while addressing the key issue of aliasing artifacts during multi-scale construction. Instead of relying purely on digital downsampling, FVAR introduces a next-focus prediction paradigm that progressively reduces blur rather than downsampling, creating smooth and alias-free transitions between scales.

\subsection{Scale-wise VAR and Architectural Variants}
VAR reformulates AR as next-scale prediction over a multi-scale token pyramid, utilizing structured masking to preserve 2D spatial structures and scale favorably \citep{Tian2024VAR}. Subsequent variants improve efficiency by decoupling intra-/inter-scale dependencies or replacing long-range attention with linear state-space modules (e.g., Mamba), while maintaining strong within-scale modeling \citep{mamba1, mamba2}. Other research explores improved tokenizers (e.g., LlamaGen, ViTVQ) and continuous-token AR to alleviate quantization limits (e.g., GIVT; MAR), though these may sacrifice inference speed \citep{Yu2024LlamaGen}. Our work, FVAR, is orthogonal and complementary to these directions. Rather than constructing multi-scale representations via purely digital downsampling—which introduces aliasing (jaggies, stair-stepping, moiré) and burdens the decoder with de-aliasing—we replace next-scale prediction with an optics-inspired next-focus prediction. Concretely, we build progressive refocusing pyramids using physics-consistent defocus kernels to create clean, alias-free views, and we combine this with high-frequency residual learning via a lightweight teacher network. This design preserves the standard single-decoder VAR pipeline and token type while substantially suppressing aliasing at its source and enhancing fine detail and text fidelity.

\subsection{Anti-Aliasing and Demoiré Removal in Visual Modelling}
Aliasing artifacts, such as jaggies and moiré patterns, arise when high-frequency image content exceeds the sampling capability (i.e., violates the Nyquist criterion), causing signal folding into lower frequencies. In rendering graphics and image processing, classic anti-aliasing techniques (e.g., supersampling, morphological anti-aliasing) mitigate jagged edges by applying pre-filtering or super-sampling before downsampling \citep{SupersamplingWiki, MLAAWiki}. In the deep-learning era, dedicated research on moiré-pattern removal (“demoiréing”) has emerged. For instance, Liu et al. propose learning in both the frequency and spatial domains via wavelet transforms and spatial demoiré blocks to separate moiré structures from image content \citep{Liu2020}. In addition, methods like Depth Adaptive Blur-Pool replace naïve downsampling to reduce aliasing in deep neural networks \citep{Hossain2021}. More recently, dual-domain multi-scale networks have been developed to address moiré removal across multiple resolutions \citep{Zhang2024}. While these techniques have primarily focused on image restoration or classification, fewer have tackled aliasing or moiré artifacts explicitly in the generation pipeline, especially in visual autoregressive models. This gap highlights the novelty of our FVAR approach, which proactively suppresses aliasing during construction—through a blur-to-clarity pyramid and residual high-frequency teacher—rather than relying on the decoder to correct these issues post-processing.

\section{Method}
Existing visual autoregressive models depend on uniform downsampling to build multi-scale representations, inevitably introducing aliasing artifacts that degrade generation quality. We address this by shifting the paradigm from next-scale to next-focus prediction grounded in optical imaging physics. FVAR contributes three core innovations: \textbf{(1) a Next-Focus Prediction Paradigm} enabling alias-free, focus-based autoregression; \textbf{(2) Progressive Refocusing Pyramid Construction} using physics-consistent defocus modeling; and \textbf{(3) High-Frequency Residual Learning} that leverages complementary high-frequency cues through teacher–student distillation while preserving full compatibility.

\subsection{Next-Focus Prediction Paradigm}
Our FVAR framework implements the next-focus prediction paradigm through three key components: progressive refocusing pyramid construction, dual-path tokenization, and high-frequency residual learning via a specialized teacher network.

We propose a paradigm shift from \emph{scale-based} to \emph{focus-based} autoregression, grounded in the physics of optical image formation. Instead of predicting increasingly downsampled versions, we model the natural focusing process where optical blur progressively decreases:
\begin{equation}
\mathcal{F}: \quad x \rightarrow \{F_{\rho_1}(x), F_{\rho_2}(x), \ldots, F_{\rho_K}(x)\},
\label{eq:focus-mapping}
\end{equation}
where \(F_{\rho_k}(x) = (k_{\rho_k} \star x)\) represents the convolution with a defocus kernel of radius \(\rho_k\), and \(\rho_1 > \rho_2 > \cdots > \rho_K = 0\).

This formulation offers several theoretical advantages: \textbf{(1) Spectral Preservation}: Each focus state \(F_{\rho_k}(x)\) is band-limited by the frequency response of point spread function (PSF), preventing aliasing artifacts. \textbf{(2) Continuity}: The focus sequence forms a continuous manifold in the space of blur kernels, enabling smooth interpolation between states. \textbf{(3) Information Monotonicity}: Information content increases monotonically as \(\rho_k \to 0\), aligning with the autoregressive generation process.

\subsection{Progressive Refocusing Pyramid Construction}
We implement the next-focus prediction through physics-consistent defocus modeling that naturally eliminates aliasing artifacts at their source, as illustrated in \cref{fig:method}. The defocus PSF for a circular aperture is approximated as a normalized disk kernel \(k_{\rho}\), where the radius follows a monotonically decreasing schedule:
\begin{equation}
\rho_k = \rho_{\max} \cdot \frac{1 - \cos\left(\pi \frac{k-1}{K-1}\right)}{2}, \quad k = 1, 2, \ldots, K,
\label{eq:focus-schedule}
\end{equation}
ensuring smooth blur-to-clarity transitions from \(\rho_1 > \rho_2 > \cdots > \rho_K = 0\).

To capture both overall structure and high-frequency residual information, we construct complementary views through our dual-path strategy:
\begin{align}
L_k &= \left(k_{\rho_k} \star x\right) \downarrow_{s_k} + \beta_k \varepsilon, \label{eq:focus-view} \\
D_k &= x \downarrow_{s_k}, \quad A_k = D_k - L_k, \label{eq:alias-residual}
\end{align}
where \(L_k\) represents the physics-consistent focused view, \(D_k\) the traditional downsampled view, and \(A_k\) the high-frequency residual information. The noise term \(\beta_k \varepsilon\) ensures full-rank covariance and training stability.

\subsection{High-Frequency Residual Learning via Specialized Teacher Network}
\begin{figure}[t]
\centering
\includegraphics[width=1.\linewidth]{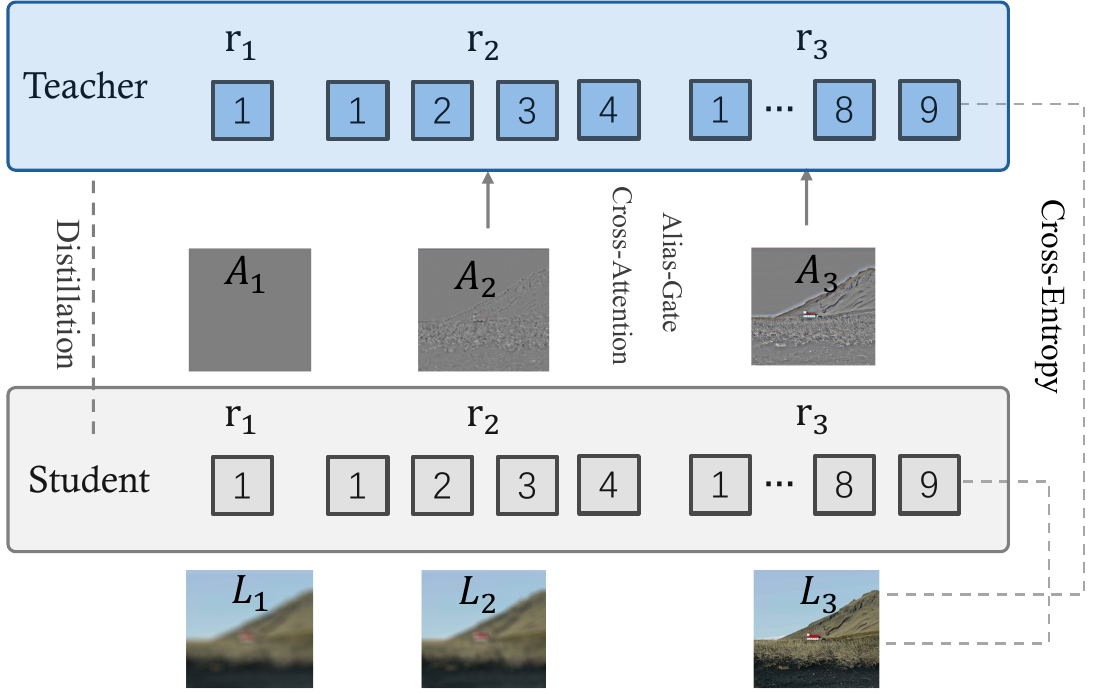}
\caption{\textbf{High-Frequency Residual Teacher Training Architecture.} Our approach employs dual networks during training: the High-Frequency Residual Teacher (top) processes both structure tokens $r_k$ and alias tokens $a_k$ through Alias-Gate Cross-Attention, while the Deployment Network (bottom) only uses structure tokens to maintain vanilla VAR compatibility. Residual knowledge transfer enables the deployment network to benefit from high-frequency information during training while ensuring zero inference overhead.}
\label{fig:architecture}
\vspace{-1.5em}
\end{figure}

While progressive refocusing pyramids provide clean, alias-free representations, the high-frequency residuals \(A_k\) contain valuable information for detail generation. To exploit these signals without altering the deployment architecture, we introduce a High-Frequency Residual Teacher that decouples alias-aware training from inference, as shown in \cref{fig:architecture}.

We tokenize both the focused views and high-frequency residuals using our dual-path strategy: \(r_k = Q_L(L_k)\) and \(a_k = Q_A(A_k)\), where the alias codebook \(|\mathcal{C}_A| \ll |\mathcal{C}_L|\) reflects the sparse nature of high-frequency patterns. During training, the High-Frequency Residual Teacher incorporates standard self-attention on structure tokens plus Alias-Gate Cross-Attention to selectively fuse information from both token streams, while the deployment network operates solely on structure tokens using standard self-attention, maintaining full compatibility with VAR.
Residual knowledge transfer passes the teacher’s enhanced capabilities to the deployment network through multi-level objectives:
\begin{equation}
\begin{split}
\mathcal{L}_{\text{total}} =\;& \mathcal{L}_{\text{AR}}^{\text{deploy}} + \lambda_{\text{feat}} \mathcal{L}_{\text{feat}}\\
&{}+ \lambda_{\text{logit}} \mathcal{L}_{\text{logit}},
\end{split}
\label{eq:total-loss}
\end{equation}
where $\mathcal{L}_{\text{feat}}$ enforces feature alignment and $\mathcal{L}_{\text{logit}}$ aligns the output distributions. During inference, only the deployment network is used, introducing no additional overhead and preserving full compatibility with the VAR framework.
The proposed FVAR integrates the three components: progressive pyramid construction generates dual-path representations, the High-Frequency Residual Teacher learns from both structure and alias tokens, and residual knowledge transfer enables vanilla VAR deployment with zero inference overhead. The overall complexity remains comparable to vanilla VAR: \textbf{(1) PSF Construction}: \(\mathcal{O}(K \cdot H \cdot W \cdot \rho_{\max}^2)\) for \(K\) focus states, which can be precomputed and cached. \textbf{(2) Teacher Training}: Additional \(\mathcal{O}(N^2 d)\) for AG-XAttn per selected layer, where \(N\) is sequence length and \(d\) is hidden dimension. With \(M \in \{1,2\}\) layers, this adds ~6-15\

\subsection{Spectral Analysis of Aliasing Decomposition}
From a signal processing perspective, uniform downsampling without anti-aliasing prefiltering causes spectral folding that maps supra-Nyquist frequencies into the baseband. For a 1D signal undergoing 2:1 decimation, the Fourier transform of the downsampled signal within the baseband \(\omega \in [-\pi/2, \pi/2]\) becomes:
\begin{equation}
\hat{D}(\omega) = \frac{1}{2}\left[X(\omega) + X(\omega + \pi)\right],
\label{eq:aliasing-1d}
\end{equation}
where \(X(\omega + \pi)\) represents the folded high-frequency content. In 2D, similar spectral folding occurs along each spatial dimension.

With an ideal anti-aliasing filter \(H_k\) having cutoff frequency \(\pi/2\), the baseband spectrum decomposes as \(D_k = L_k + A_k\), where the alias residual in the frequency domain satisfies:
\begin{equation}
\hat{A}_k(\omega) = \frac{1}{2} \sum_{u \in \mathcal{U}} X(\omega + u),
\label{eq:alias-spectrum}
\end{equation}
with \(\mathcal{U}\) denoting the set of folding shift vectors per spatial axis. This decomposition has the following properties:

\paragraph{Alias-free structure preservation.} If \(H_k\) implements ideal low-pass filtering with cutoff \(\pi/2\), then \(\hat{L}_k(\omega) = X(\omega)\) for \(|\omega| \leq \pi/2\), ensuring the low-frequency view \(L_k\) contains no aliasing artifacts within the passband.

\paragraph{Predictive high-frequency evidence.} The alias residual \(A_k\) aggregates folded high-frequency content that encodes valuable information about edge orientations, texture patterns, and fine-scale structures, making it a complementary signal for detail recovery.

\paragraph{Energy conservation.} The spectral energy of the alias residual satisfies:
\begin{equation}
\|\hat{L}_k - \hat{D}_k\|_2^2 = \|\hat{A}_k\|_2^2 = \frac{1}{4}\|X(\omega + \pi)\|_2^2
\label{eq:alias-energy}
\end{equation}
within the passband, providing direct control over aliasing through the choice of \(H_k\).

From an optimization standpoint, VQ codebooks trained on \(L_k\) operate on smoother, better-conditioned signals with improved numerical stability, while the alias cues in \(a_k\) can be selectively incorporated when helpful for enhancing fine details.

\subsection{Alias-Gate Cross-Attention in Teacher Network}
To enable the teacher network to exploit high-frequency alias information during training, we introduce \emph{Alias-Gate Cross-Attention} (AG-XAttn), a lightweight module applied only in the teacher’s encoder. Importantly, the student network omits AG-XAttn, preserving the vanilla VAR architecture for full deployment compatibility. In the teacher’s encoder blocks—applied selectively to the final $M$ autoregressive scales for efficiency—we first compute windowed self-attention on the structure tokens, followed by cross-attention from structure to alias:
\begin{equation}
\label{eq:agxattn}
\begin{aligned}
Z &= \text{WSA}\bigl(E(r_k)\bigr) + \text{Attn}\Bigl(
      \begin{aligned}[t]
        Q &= X_L W_Q, \\
        K &= E_a(a_k) W_K,\; V = E_a(a_k) W_V
      \end{aligned}
    \Bigr),
\end{aligned}
\end{equation}
where \(E(\cdot)\) and \(E_a(\cdot)\) denote the structure and alias token embeddings, respectively, and \(W_Q, W_K, W_V \in \mathbb{R}^{d \times d_h}\) are learned projection matrices. The resulting contextual representations \(C_k = Z\) are fed to the unchanged decoder, while the alias tokens \(\{a_k\}\) remain excluded from the autoregressive prediction sequence.

\paragraph{Wiener filtering interpretation.} Under local linearization, the cross-attention update can be viewed as a learned gated residual connection:
\begin{equation}
Z \approx X_L + \alpha \odot \tilde{A}_k,
\label{eq:wiener-analogy}
\end{equation}
where \(\alpha \in [0,1]^d\) represents a data-dependent gating function and \(\tilde{A}_k\) denotes the processed alias information. This resembles the classical Wiener filter formulation, where the optimal gain for MSE minimization is:
\begin{equation}
\alpha^*(\omega) = \frac{S_{xx}(\omega)}{S_{xx}(\omega) + S_{nn}(\omega)},
\label{eq:wiener-gain}
\end{equation}
with \(S_{xx}(\omega)\) and \(S_{nn}(\omega)\) representing the signal and noise power spectral densities, respectively. Intuitively, the learned attention mechanism adaptively upweights reliable, edge-aligned frequencies while suppressing aliasing patterns prone to generating moiré artifacts.

\begin{table*}[t]
\renewcommand\arraystretch{1.05}
\centering
\setlength{\tabcolsep}{2.5mm}
\small
\caption{\textbf{Comparisons on ImageNet 256$\times$256, 512$\times$512, and 1024$\times$1024}. Metrics: FID$\downarrow$, IS$\uparrow$, Precision (Pre)$\uparrow$, Recall (Rec)$\uparrow$. Step: model runs to generate one image. Time: relative inference time normalized to VAR-d12 at 256$\times$256 and VAR-d16 at higher resolutions (Time=1.0).}
\vspace{-1.em}
\label{tab:main}
\begin{tabular}{l|cc|cc|cc|c}
\toprule
 Model          & FID$\downarrow$ & IS$\uparrow$ & Pre$\uparrow$ & Rec$\uparrow$ & Param & Step & Time \\
\midrule
\multicolumn{8}{c}{\emph{Generative Adversarial Net (GAN)}} \\
 BigGAN~\citep{biggan}  & 6.95  & 224.5       & 0.89 & 0.38 & 112M & 1    & --    \\
 GigaGAN~\citep{gigagan}     & 3.45  & 225.5       & 0.84 & 0.61 & 569M & 1    & -- \\
 StyleGAN-XL~\citep{sauer2022stylegan}  & 2.30  & 265.1       & 0.78 & 0.53 & 166M & 1    & 0.2 \\
\midrule
\multicolumn{8}{c}{\emph{Diffusion}} \\
 ADM~\citep{adm}         & 10.94 & 101.0        & 0.69 & 0.63 & 554M & 250  & 118 \\
 CDM~\citep{cdm}         & 4.88  & 158.7       & --  & --  & --  & 8100 & --    \\
 LDM-4-G~\citep{ldm}     & 3.60  & 247.7       & --  & --  & 400M & 250  & --    \\
 DiT-L/2~\citep{dit}     & 5.02  & 167.2       & 0.75 & 0.57 & 458M & 250  & 2     \\
 DiT-XL/2~\citep{dit}    & 2.27  & 278.2       & 0.83 & 0.57 & 675M & 250  & 2     \\
 L-DiT-7B~\citep{dit-github}    & 2.28  & 316.2       & 0.83 & 0.58 & 7.0B & 250  & $>$32     \\
\midrule
\multicolumn{8}{c}{\emph{Mask Prediction}} \\
 MaskGIT~\citep{maskgit}     & 6.18  & 182.1        & 0.80 & 0.51 & 227M & 8    & 0.4  \\
 RCG (cond.)~\citep{rcg}  & 3.49  & 215.5        & --  & --  & 502M & 20  & 1.4  \\
\midrule
\multicolumn{8}{c}{\emph{Token-wise Autoregressive}} \\
 VQGAN~\citep{vqgan}       & 15.78 & 74.3   & --  & --  & 1.4B & 256  & 17     \\
 ViTVQ~\citep{vit-vqgan}   & 4.17  & 175.1  & --  & --  & 1.7B & 1024 & $>$17  \\
 RQTran.~\citep{rq}        & 7.55  & 134.0  & --  & --  & 3.8B & 68   & 15     \\
 LlamaGen-3B~\citep{llamagen}& 2.18& 263.3 &0.81& 0.58 &3.1B&576& --\\
\midrule
\multicolumn{8}{c}{\emph{Scale-wise Autoregressive}} \\
 VAR-d12~\citep{var}         & 5.81 & 201.3 & 0.81 & 0.45 & 132M & 10   & 1.0  \\
M-VAR-d12~\citep{mvar}      & 4.19 & 234.8 & 0.83 & 0.48 & 198M & 10   & 1.0  \\
\rowcolor{gray!20}
FVAR-d12 (Ours)      & \textbf{3.95} & \textbf{238.2} & \textbf{0.84} & \textbf{0.49} & 132M & 10 & 1.0  \\
VAR-d16~\citep{var}         & 3.55 & 280.4 & 0.84 & 0.51 & 310M & 10   & 1.0  \\
M-VAR-d16~\citep{mvar}      & 3.07 & 294.6 & 0.84 & 0.53 & 464M & 10   & 1.0  \\
\rowcolor{gray!20}
FVAR-d16 (Ours)      & \textbf{2.89} & \textbf{298.1} & \textbf{0.85} & \textbf{0.54} & 310M & 10 & 1.0  \\
VAR-d20~\citep{var}         & 2.95 & 302.6 & 0.83 & 0.56 & 600M & 10   & 1.5  \\
M-VAR-d20~\citep{mvar}      & 2.41 & 308.4 & 0.85 & 0.58 & 900M & 10   & 2.0  \\
\rowcolor{gray!20}
FVAR-d20 (Ours)      & \textbf{2.25} & \textbf{312.8} & \textbf{0.86} & \textbf{0.59} & 600M & 10 & 1.5  \\
VAR-d24~\citep{var}         & 2.33 & 312.9 & 0.82 & 0.59 & 1.0B & 10   & 2.5  \\
M-VAR-d24~\citep{mvar}      & 1.93 & 320.7 & 0.83 & 0.59 & 1.5B & 10   & 3.0  \\
\rowcolor{gray!20}
FVAR-d24 (Ours)      & \textbf{1.75} & \textbf{325.8} & \textbf{0.84} & \textbf{0.61} & 1.0B & 10 & 2.5  \\
\bottomrule
\end{tabular}
\vspace{-1.em}
\end{table*}

\subsection{Teacher-Student Knowledge Distillation}
A core component of our method is the online distillation between the teacher network (equipped with AG-XAttn) and the student network (using the vanilla VAR architecture). During training, both networks process the same input batch in parallel, and knowledge transfer is facilitated through a set of complementary objectives:

\paragraph{Training Objective.} For each scale \(k\), the combined loss function is:
\begin{equation}
\begin{split}
\mathcal{L}_{\text{total}} =\;& \mathcal{L}_{\text{AR}}^{\text{stu}}\!\left(r_{k-1}, p_{\text{stu}}\right) \\
&{}+ \lambda_{\text{feat}} \sum_{\ell} \bigl\|F_{\text{stu}}^{(\ell)} - \text{sg}\!\left(F_{\text{tea}}^{(\ell)}\right)\bigr\|_2^2 \\
&{}+ \lambda_{\text{logit}} \cdot \text{KL}\!\left(p_{\text{tea}} \| p_{\text{stu}}\right),
\end{split}
\label{eq:distillation-loss}
\end{equation}
where \(\mathcal{L}_{\text{AR}}^{\text{stu}}\) is the standard autoregressive loss for the student, \(F^{(\ell)}\) denotes feature representations from the final 1-2 encoder blocks, \(\text{sg}(\cdot)\) is the stop-gradient operator, and \(p_{\text{tea}}, p_{\text{stu}}\) are the output logits from teacher and student networks respectively.

\paragraph{Deployment Strategy.} During inference, only the student network is employed, functioning exactly like a vanilla VAR model and ensuring full compatibility. The teacher network only works in training-time and is discarded when the training is completed.
\section{Experiments}

\begin{table}[t]
\centering
\small
\renewcommand\arraystretch{1.1}
\caption{\textbf{Comparisons on ImageNet 512$\times$512 and 1024$\times$1024}.}
\label{tab:highres}
\renewcommand\arraystretch{1.1}
 \setlength{\tabcolsep}{3.4mm}{
 \resizebox{1.0\linewidth}{!}{
\begin{tabular}{l|cc|cc}
\toprule
Model & \multicolumn{2}{c|}{512$\times$512} & \multicolumn{2}{c}{1024$\times$1024} \\
 & FID$\downarrow$ & IS$\uparrow$ & FID$\downarrow$ & IS$\uparrow$ \\
\midrule
VAR-d36~\citep{var}         & 2.63 & 303.2 & -- & -- \\
\rowcolor{gray!20}
FVAR-d36 (Ours)      & \textbf{2.28} & \textbf{315.6} & -- & -- \\
\midrule
VAR-d16~\citep{var}         & -- & -- & 8.25 & 298.3 \\
\rowcolor{gray!20}
FVAR-d16 (Ours)      & -- & -- & \textbf{6.85} & \textbf{315.2} \\
\bottomrule
\end{tabular}
}}
\vspace{-1.em}
\end{table}
\subsection{Datasets and Metrics}
We evaluate our method on class-conditional generation for ImageNet at $256\times256$, $512\times512$, and $1024\times1024$ resolutions, following prior VAR studies \citep{deng2009imagenet,var}. Generation quality is assessed using standard metrics, including FID \citep{NIPS2017_8a1d6947}, IS \citep{NIPS2016_8a3363ab}, and Precision/Recall \citep{NEURIPS2019_0234c510}.
\subsection{Implementation Details}
We follow the training setup of VAR \citep{var}, with adjustments to accommodate our dual-path architecture. All models are trained on $8\times$ A100 GPUs using mixed precision. For the progressive refocusing pyramid, we employ $K=4$ scales with a maximum PSF radius of $\rho_{\max}=12$ pixels and cosine scheduling. The structure codebook contains 8192 entries, while the alias codebook uses 512 entries to capture the sparse characteristics of high-frequency patterns.

The High-Frequency Residual Teacher applies AG-XAttn only to the final two transformer blocks to reduce computational cost. Knowledge distillation uses $\lambda_{\text{feat}}=1.0$ and $\lambda_{\text{logit}}=0.5$. Training is performed in two stages: first, the dual VQ tokenizers are trained for 100K steps, followed by end-to-end training for 400K steps with a learning rate of $1\!\times\!10^{-4}$ and a batch size of 256. The noise regularization $\beta_k$ increases linearly from $1\!\times\!10^{-3}$ to $1\!\times\!10^{-2}$ across scales. In addition to ImageNet, we also train our base models on additional large-scale datasets to further validate the effectiveness of our method. Detailed information about the additional training data, training procedures, and computational complexity analysis is provided in the supplementary material.
\subsection{Main results}
\begin{figure*}[t]
\centering
\includegraphics[width=0.95\textwidth]{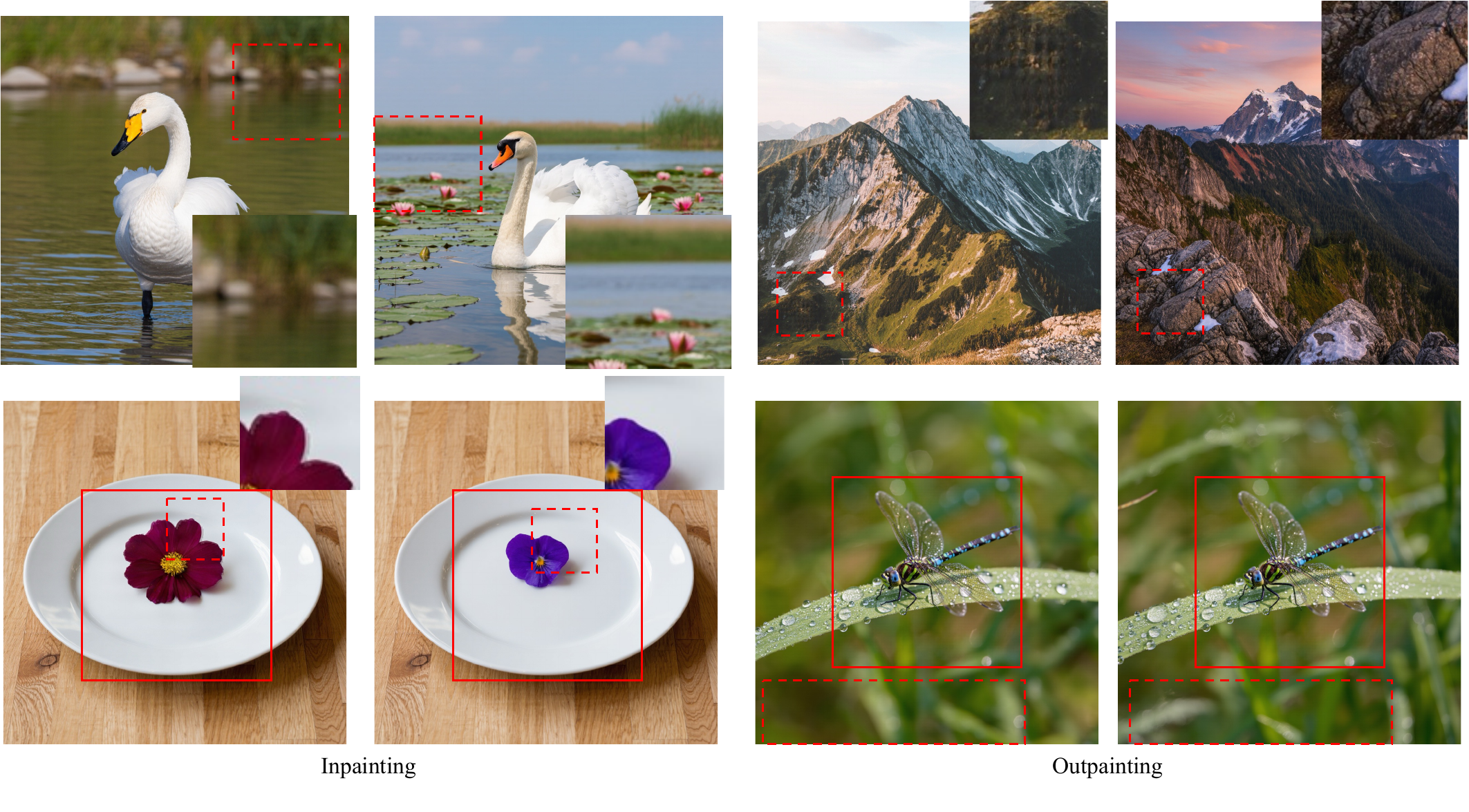}
\caption{\textbf{Visual quality comparison between VAR and FVAR.} To compare the quality of spatial hierarchy and high-frequency details, these are visualization results at 1024$\times$1024 resolution. The first row shows image generation, and the second row shows inpainting and outpainting (solid red boxes indicate input regions). In each group, VAR is on the left and FVAR is on the right. Dashed red boxes highlight key regions of interest.}
\vspace{-1.5em}
\label{fig:visual_comparison}
\end{figure*}

Table~\ref{tab:main} shows our method consistently outperforms both VAR and M-VAR across different model sizes at 256×256 resolution, achieving better FID scores with comparable inference speed. Table~\ref{tab:highres} demonstrates consistent improvements at higher resolutions (512×512 and 1024×1024), validating the scalability of our method. Figure~\ref{fig:visual_comparison} demonstrates that FVAR significantly reduces aliasing artifacts while preserving fine details. Specifically, in the top-left comparison, our method handles the highlighted region more naturally without high-frequency texture artifacts, while the top-right comparison reveals obvious grid-like jaggies in VAR. The bottom-left comparison shows aliasing artifacts in VAR's inpainting results, whereas the bottom-right demonstrates better spatial hierarchy in our outpainting generation.

\subsection{Ablations and analysis}
We perform ablation studies to evaluate the contribution of each component of our method. Unless stated otherwise, all experiments use the VAR-d16 architecture on ImageNet $256\times256$. Detailed results are provided in Table~\ref{tab:ablation}. To further validate the effectiveness of our method, we conduct additional ablation studies at 1024×1024 resolution, as shown in Table~\ref{tab:ablation_highres}. The results demonstrate that removing key components leads to significant performance degradation at higher resolutions. This observation highlights that our method becomes increasingly beneficial as resolution increases, where aliasing artifacts are more pronounced and harder to correct through post-processing.

\begin{table}[t]
\renewcommand\arraystretch{1.1}
\caption{\textbf{Ablation study on FVAR-d16.} All metrics evaluated on ImageNet 256$\times$256.}
\label{tab:ablation}
\centering
\small
\renewcommand\arraystretch{1.1}
 \setlength{\tabcolsep}{5.4mm}{
 \resizebox{1.0\linewidth}{!}{
\begin{tabular}{l|cc}
\toprule
Variant & FID$\downarrow$ & IS$\uparrow$ \\
\midrule
VAR-d16 (Baseline) & 3.55 & 280.4 \\
\rowcolor{gray!20}
FVAR-d16 (Full) & \textbf{2.89} & \textbf{298.1} \\
\midrule
w/o Progressive Refocusing & 3.51 & 282.1 \\
w/ Gaussian blur & 3.32 & 286.7 \\
w/o High-Freq Teacher & 3.06 & 294.8 \\
w/o Dual tokenizers & 3.14 & 292.1 \\
\bottomrule
\end{tabular}}}
\vspace{-.5em}
\end{table}

\begin{table}[t]
\caption{\textbf{Ablation study on FVAR-d16 at 1024$\times$1024.} All metrics evaluated on ImageNet 1024$\times$1024.}
\label{tab:ablation_highres}
\centering
\small
\renewcommand\arraystretch{1.1}
 \setlength{\tabcolsep}{5.4mm}{
 \resizebox{1.0\linewidth}{!}{
\begin{tabular}{l|cc}
\toprule
Variant & FID$\downarrow$ & IS$\uparrow$ \\
\midrule
VAR-d16 (Baseline) & 8.25 & 298.3 \\
\rowcolor{gray!20}
FVAR-d16 (Full) & \textbf{6.85} & \textbf{315.2} \\
\midrule
w/o Progressive Refocusing & 8.15 & 299.0 \\
w/ Gaussian blur & 7.50 & 305.2 \\
w/o High-Freq Teacher & 7.20 & 308.5 \\
w/o Dual tokenizers & 7.40 & 306.8 \\
\bottomrule
\end{tabular}}}
\vspace{-1.em}
\end{table}

\paragraph{Progressive Refocusing.} Progressive refocusing is a critical component of our method. At 256×256, while removing progressive refocusing shows minimal FID improvement over the baseline, this does not reflect its true impact on image quality. Standard metrics like FID primarily capture overall distribution statistics and may not adequately reflect high-frequency detail preservation and aliasing artifacts, which are better observed through visual inspection. At 1024×1024, removing this component causes severe performance degradation approaching VAR's baseline in both metrics and visual quality, clearly demonstrating its critical importance at higher resolutions. Even simple Gaussian blur provides meaningful gains compared to standard downsampling, and our physics-consistent PSF achieves substantially better results, demonstrating that optical realism in defocus modeling is crucial. The resolution-dependent behavior highlights that progressive refocusing addresses fundamental limitations that become increasingly problematic as resolution scales.

\paragraph{High-Frequency Residual Teacher.} The High-Frequency Residual Teacher plays an important role in enhancing detail generation quality. The teacher network with its specialized AG-XAttn mechanism effectively captures and transfers high-frequency information to the deployment network, confirming that our teacher-student framework substantially improves fine detail preservation. While this component shows moderate impact at lower resolutions, its contribution becomes more significant at higher resolutions where fine details are more critical.

\vspace{-2mm}
\paragraph{Dual-Path Strategy Validation.} The dual tokenizer approach provides substantial improvement over using shared tokenizers, validating that specialized quantization for different signal types is essential. This confirms our hypothesis that structure and alias information have fundamentally different statistical properties requiring separate codebook designs optimized for their respective characteristics.

\section{Conclusion}
In this work, we introduced FVAR, a physics-driven reformulation of visual autoregressive modeling that replaces conventional next-scale prediction with a novel next-focus prediction paradigm. By constructing multi-scale representations through progressive optical refocusing, FVAR fundamentally eliminates aliasing at its origin and provides smoother, more coherent transitions across scales. Combined with our dual-path tokenization and the High-Frequency Residual Teacher, the model learns to leverage both clean structural cues and informative alias residuals, ultimately distilling this capability into a fully VAR-compatible deployment network. Extensive experiments across multiple resolutions demonstrate that FVAR consistently improves fine-detail fidelity, text clarity, and overall perceptual realism without introducing any inference overhead. We believe this work opens a new direction for physics-informed autoregressive modeling, and future exploration of richer optical priors or joint camera-model learning may further advance the quality and controllability of visual generative models.

\vspace{4pt} \noindent \textbf{Limitation.} Despite its advantages, FVAR has several limitations. Mismatches in the PSF shape or radius can degrade the statistical quality of the high-frequency residuals $A_k$. The dual-codebook training scheme may require careful calibration, such as embedding sharing or distillation into a single codebook, to ensure seamless deployment. Additionally, extremely high-frequency text patterns (less than 2 pixels) may still pose challenges and could require additional textual priors or specialized correction heads to preserve fine details.

{\small
\bibliographystyle{ieeenat_fullname}
\bibliography{main}
}
\end{document}